\documentclass[lettersize,journal]{IEEEtran}
\usepackage{amsmath,amsfonts}
\usepackage{algorithmic}
\usepackage{algorithm}
\usepackage{amssymb}
\usepackage{array}
\usepackage[caption=false,font=normalsize,labelfont=sf,textfont=sf]{subfig}
\usepackage{textcomp}
\usepackage{stfloats}
\usepackage{url}
\usepackage{verbatim}
\usepackage{graphicx}
\usepackage{cite}
\usepackage{hyperref}
\hypersetup{colorlinks=true, linkcolor=blue, citecolor=blue, urlcolor=blue}
\newcommand{\figref}[2]{\hyperref[#1]{Fig.~\ref*{#1}#2}}

\usepackage{tabularx}
\newcolumntype{C}{>{\centering\arraybackslash}X}
\usepackage{booktabs}
\usepackage{adjustbox}
\usepackage{xcolor} % 用于红色/蓝色高亮
\usepackage{multirow}
\usepackage{makecell}
\usepackage{float}

\begin{document}

\title{Degradation-Aware Adaptive Context Gating for Unified \\Image Restoration}

\author{
Lei He,
Jielei Chu,~\IEEEmembership{Senior~Member,~IEEE,}
Fengmao Lv,~\IEEEmembership{Member,~IEEE,}
Weide Liu,~\IEEEmembership{Member,~IEEE,}
Tianrui Li,~\IEEEmembership{Senior~Member,~IEEE,}
Jun Cheng,~\IEEEmembership{Senior~Member,~IEEE,}
Yuming Fang,~\IEEEmembership{Fellow,~IEEE,}
        % <-this % stops a space
\thanks{Lei He is with the School of Computing and Artificial Intelligence, Southwest Jiaotong University, Chengdu 611756, China. e-mail: (2024212445@my.swjtu.edu.cn).

Jun Cheng is with the Institute for Infocomm Research, Agency for Science,
Technology and Research, Singapore 138632. (e-mail: juncheng@ieee.org).

Weide Liu is with School of Computing and Artificial Intelligence, Jiangxi University of Finance and Economics, Nanchang, 330013, China. (weide001@e.ntu.edu.sg).

Yuming Fang is with the School of Information Management, Jiangxi University of Finance and Economics, Nanchang, China. (e-mail: fa0001ng@e.ntu.edu.sg).

Jielei Chu, Fengmao Lv, Tianrui Li are with the School of Computing
and Artificial Intelligence, Southwest Jiaotong University, Chengdu 611756,
China. Jielei Chu is also with the Engineering Research Center of Sustainable Urban
Intelligent Transportation, Ministry of Education, Chengdu, Sichuan
611756, China. Tianrui
Li is also with National Engineering Laboratory of Integrated Transportation
Big Data Application Technology, Southwest Jiaotong University, Chengdu
611756, China. (e-mail: \{fengmaolv, jieleichu, trli\}@swjtu.edu.cn).
\vspace{1em} % 在这里插入垂直间距，数值可根据需要调整，如 1em, 10pt 等

Jielei Chu is the corresponding author. The paper has been submitted to the IEEE TIP journal.}
}

\maketitle

\begin{abstract}
Unified image restoration aims to handle diverse degradation types using a single model. However, the significant variability across different degradations often leads to severe task interference and suboptimal performance. Existing methods often struggle to balance task-specific discriminability with inter-task generalization, leading to either negative interference in complex environments or suboptimal performance on specific degradations. To overcome these challenges, we propose a Degradation-Aware Adaptive Context Gating (DACG-IR), which enables the restoration model to explicitly perceive degradation characteristics and dynamically modulate feature representations conditioned on the input image. The core idea is to construct degradation-aware contextual representations directly from the input image and utilize them to modulate attention distribution, frequency-domain modulation, and feature aggregation throughout the model. This design enables the model to suppress degradation-induced noise and interference while preserving informative image structures. Specifically, we design a lightweight multi-scale degradation-aware module to extract coarse degradation information and generate layer-wise degradation prompts, which guide the attention temperature and attention output gating in different blocks of the encoder and decoder, enabling adaptive feature extraction and fusion across scales. The generated global feature prompts is further used to dynamically modulate high-dimensional latent features. Furthermore, a spatial–channel dual-gated adaptive fusion mechanism is designed to refine encoder features and suppress the propagation of noise or irrelevant background information from shallow layers to deeper representations, thereby promoting high-fidelity reconstruction in the decoder. Extensive experiments on multiple benchmark datasets show that DACG-IR consistently outperforms state-of-the-art image restoration methods under single-task, all-in-one, adverse weather removal, and composite degradation settings. \noindent\textbf{Code:} \url{https://github.com/HlHomes/DACG-IR-code}.
\\
\end{abstract}

\section{Introduction}
\IEEEPARstart{I}{mage} restoration focuses on reconstructing high-quality visual content from images affected by diverse degradation processes~\cite{TPAMI2025survey}, \cite{Snow100K_DesnowNet}, \cite{CNN_HDCW-Net}, \cite{cvpr2022Uformer},\cite{Dehazefomer}. 
In real-world scenarios, images are often affected by complex degradations caused by imaging devices or adverse environments, such as haze, noise, rain streaks, blur, and low-light conditions, which may occur simultaneously. 
Image restoration plays a crucial role in high-level vision tasks, including image recognition~\cite{image_recognition}, object detection~\cite{object},  and visual tracking~\cite{cvpr2022_object_track}, by providing reliable visual representations. 
Early restoration approaches mainly focus on specific degradation types and rely on physical models or hand-crafted priors~\cite{Rain100L},\cite{SOTS},\cite{CBSD68}. 
Although effective under certain assumptions, these task-specific methods lack the flexibility to handle unknown or mixed degradations. 
The performance of image restoration techniques has been significantly advanced by recent progress in deep neural networks, yet most of them remain specialized for individual restoration tasks \cite{Dehazefomer},\cite{cvpr2023denoising}, \cite{LOL_Retinex-Net}, \cite{DeblurGAN},\cite{SPAData}.
However, the unpredictable nature of real-world degradations makes model training challenging and increases computational costs.

\begin{figure}[t]
\centering
\includegraphics[width=\columnwidth]{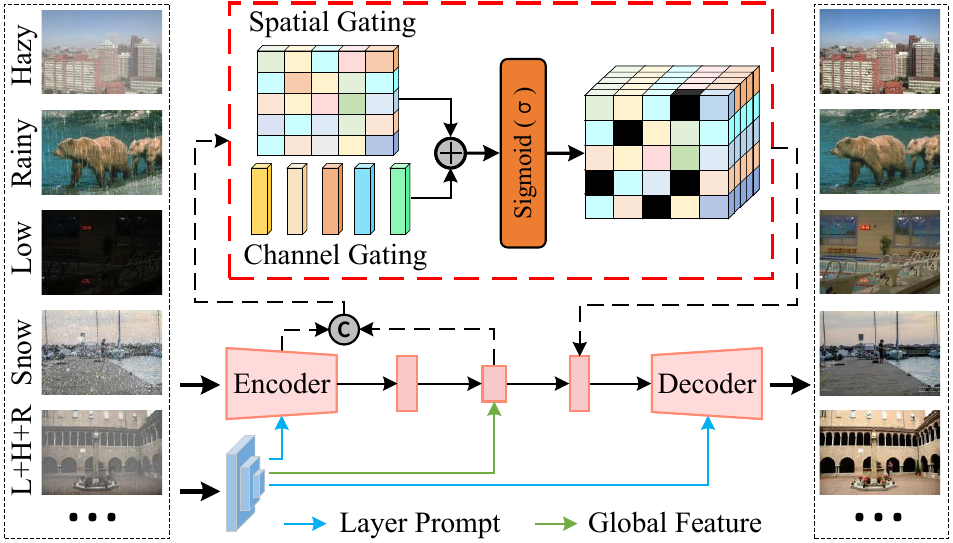}
\caption{The proposed Degradation-aware Adaptive Context Gating framework for unified image restoration. layer prompts guide the attention mechanism, while global feature perform frequency-domain modulation. A dual-path gating mechanism in the spatial and channel domains adaptively filters degradation noise, enabling robust restoration under diverse adverse conditions.}
\label{fig:motivation}
\end{figure}

To address the limitations of single-task restoration, in recent years, unified frameworks for image restoration have been introduced to simultaneously tackle a variety of degradation types within a single architecture\cite{tip2025m2restore}, \cite{tip2025Perceive_IR}, \cite{cvpr2025MoCE_IR}, \cite{DFPIR}, \cite{DynamicPrompt}, \cite{DA-RCOT}, \cite{cvpr2022transweather}, \cite{FSNet-TPAMI}, \cite{PromptIR}.
Unified restoration offers a more effective and practical solution for real-world scenarios. 
For example, AirNet~\cite{cvpr2022AirNet} achieves unified restoration through contrastive degradation learning, while PromptIR~\cite{PromptIR} introduces adaptive prompts to encode degradation information. 
Subsequent works further enhance restoration by distinguishing degradation characteristics through feature perturbation~\cite{DFPIR} or mixture-of-experts architectures~\cite{cvpr2025MoCE_IR}. 
Despite promising results, unified restoration remains challenging because different degradations exhibit heterogeneous physical properties and spatial–frequency characteristics. 
For instance, noise~\cite{CBSD68},\cite{cvpr2023denoising},\cite{Urban100}, and low-light conditions~\cite{LOL_Retinex-Net},\cite{cvpr2025darkir}, mainly affect local intensity statistics, whereas haze~\cite{DehazeNet} and rain~\cite{RainDrop} introduce global contrast degradation and structured artifacts. 
Most existing methods rely on large-capacity networks to implicitly learn degradation-invariant representations. 
However, they lack explicit mechanisms to separate degradation-related information from useful image structures. 
Consequently, early-layer features may contain degradation artifacts that propagate through the network, especially in encoder–decoder architectures with skip connections\cite{cvpr2022Uformer},~\cite{NAFNet},\cite{cvpr_2022_Restormer}, which can negatively affect reconstruction quality.

\begin{figure}[t]
\centering
\includegraphics[width=\columnwidth]{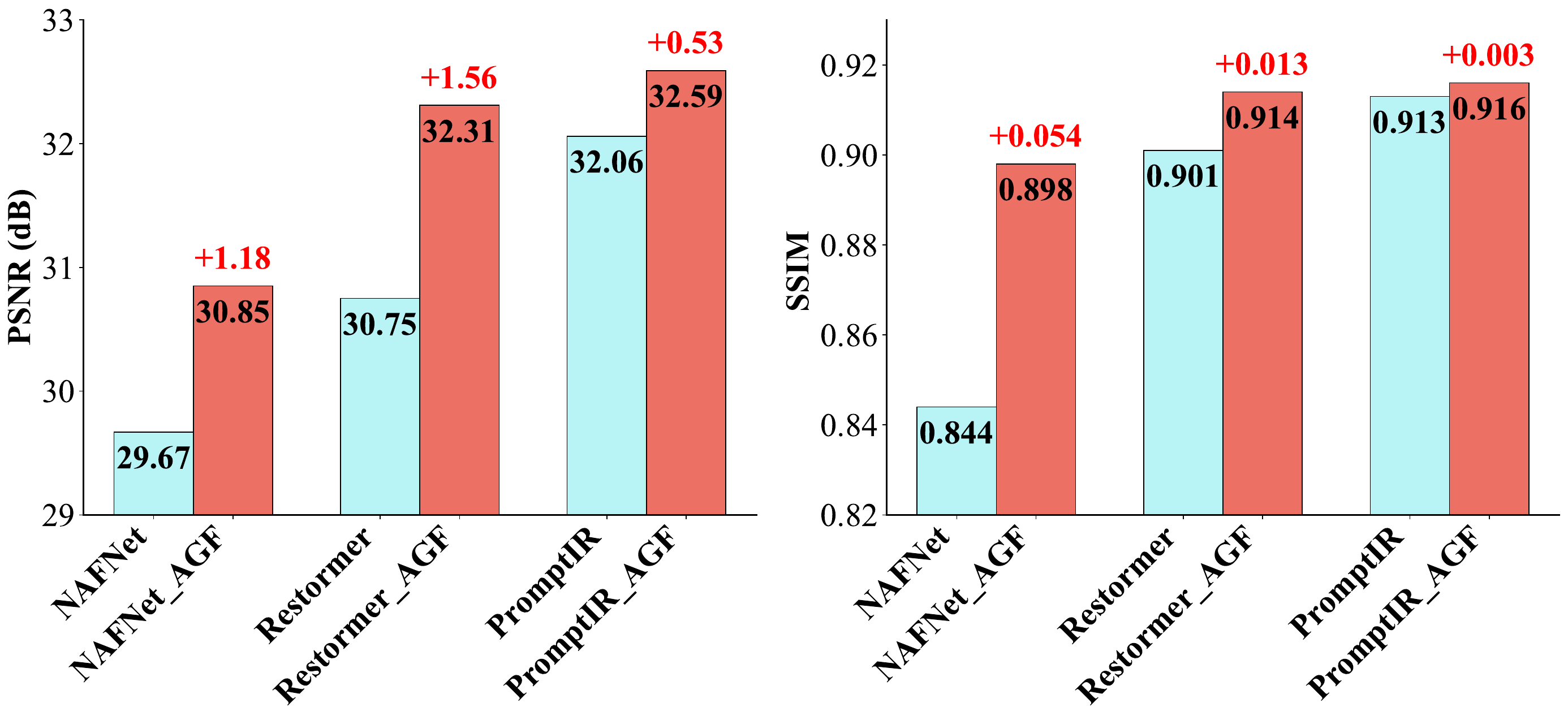}
\caption{Average results on the three-task all-in-one image restoration benchmark. By incorporating the proposed AGF module into three representative architectures(NAFNet\cite{NAFNet}, Restormer\cite{cvpr_2022_Restormer}, and PromptIR\cite{PromptIR}), we observe consistent and architecture-agnostic improvements in both PSNR and SSIM. This suggests that AGF serves as a general and effective feature fusion strategy for unified image restoration under diverse degradations.}
\label{fig:AGF_Comparison_Three_Task}
\end{figure}

The key challenge of unified image restoration lies in adaptively regulating feature extraction and fusion according to latent degradation characteristics without relying on explicit degradation labels. 
An effective unified model should: 
1) perceive degradation information directly from the input image; 
2) enhance informative structures while suppressing degradation interference; and 
3) adaptively control feature propagation across network layers. 
However, most existing approaches~\cite{cvpr2022Uformer},\cite{PromptIR},\cite{cvpr2022AirNet},\cite{cvpr_2022_Restormer}, treat features uniformly and adopt static attention or fusion strategies, which cannot fully accommodate the diversity of real-world degradations. 
Therefore, explicitly incorporating degradation-aware contextual information into feature modulation is crucial. 
In particular, degradation information should operate at both global and local levels: global context describes the overall degradation pattern, while layer-wise modulation prevents noisy features from propagating to deeper stages.

To better handle diverse degradations, we propose a degradation-aware adaptive context gating framework for unified image restoration, termed DACG-IR.
The proposed framework generates degradation-aware contextual prompts to guide attention responses and feature modulation across network layers. 
Meanwhile, a gating mechanism performs fine-grained feature filtering to preserve informative structures while suppressing degradation-specific noise. 
Furthermore, we introduce an adaptive gated fusion (AGF) module to replace conventional skip connections, enabling spatial–channel feature modulation before encoder features are propagated to the decoder. 
This mechanism effectively prevents shallow noisy features from contaminating deeper representations (Fig.~\ref{fig:motivation}). 
Importantly, the gating strategy is fully data-driven and does not require explicit degradation annotations, allowing DACG-IR to flexibly adapt to diverse degradations. 
Moreover, the AGF module can be seamlessly integrated into standard encoder–decoder architectures with minimal computational overhead. 
We incorporate AGF into representative backbones including NAFNet~\cite{NAFNet}, Restormer~\cite{cvpr_2022_Restormer}, and PromptIR~\cite{PromptIR}, and observe consistent performance improvements under a unified three-task setting (Fig.~\ref{fig:AGF_Comparison_Three_Task}).

The main contributions of our work are summarized as follows:
\begin{itemize}
\item We propose a Degradation-Aware Adaptive Context Gating framework for unified image restoration, enabling robust restoration across diverse degradation scenarios.
\item We design a spatial–channel dual-gated mechanism that adaptively filters noisy features across network layers and can be seamlessly integrated into different architectures.
\item We propose a degradation-aware context adaptive gating attention mechanism and a context-gating dual-domain modulation strategy. The attention mechanism uses degradation information to adaptively adjust the attention temperature and suppress degradation noise, while the dual-domain modulation further refines frequency-domain representations, enabling clearer degradation-aware features and more accurate image reconstruction.
\item Experimental results on multiple benchmark datasets demonstrate that DACG-IR delivers state-of-the-art or competitive restoration performance.
\end{itemize}

\section{Related work}
\subsection{Image Restoration}

\textit{Single-Degradation Image Restoration:} Image restoration targeting specific degradation types has been extensively investigated in low-level vision, with existing models typically tailored to handle a single distortion, such as 
deblurring~\cite{DeblurGAN},\cite{Transformer_tsai2022stripformer},\cite{HI-Diff},\cite{MPRNet}, denoising~\cite{cvpr2023denoising},\cite{NAFNet},\cite{InstructIR}, dehazing\cite{Dehazefomer},~\cite{SOTS},\cite{2018_GFN},  and deraining\cite{SPAData},~\cite{RainDrop},\cite{DRSFormer},\cite{MSPFN}.
Early approaches mitigated the ill-posed nature of restoration problems through hand-crafted priors and physics-based constraints~\cite{Rain100L},\cite{CBSD68}, but their limited representational capacity restricted performance in real-world scenarios. With the rapid progress and development of deep learning, data-driven methods have significantly advanced restoration quality by learning degradation-specific representations from large-scale datasets. 
CNN-based models\cite{Snow100K_DesnowNet}, \cite{CNN_HDCW-Net},\cite{DehazeNet}, \cite{NAFNet}, \cite{MSPFN}, \cite{DRBN_low_light}, effectively capture hierarchical local structures, while Transformer-based architectures\cite{cvpr2023denoising}, \cite{DA-RCOT}, \cite{cvpr2022transweather}, \cite{FSNet-TPAMI}, \cite{cvpr_2022_Restormer},  \cite{Transformer_tsai2022stripformer},\cite{DRSFormer},\cite{TIP2024MWformer}, further enhance restoration by modeling long-range dependencies. 
Representative models such as NAFNet~\cite{NAFNet} employ lightweight gating mechanisms to simplify network design, whereas Restormer~\cite{cvpr_2022_Restormer} improves efficiency through channel-wise self-attention. 
Although impressive results have been documented for specific restoration tasks, the majority of current approaches presuppose prior knowledge of the degradation type. This assumption constrains their efficacy in real-world applications, where degradations are frequently complex, unidentified, or spatially non-uniform.

\textit{All-in-One Image Restoration:}
To overcome the constraints inherent in task-specific architectures, all-in-one image restoration frameworks are designed to handle diverse degradation types within a single, unified model\cite{cvpr2025MoCE_IR}, \cite{DFPIR}, \cite{DynamicPrompt}, \cite{DA-RCOT}, \cite{cvpr2022transweather}, \cite{FSNet-TPAMI}, \cite{PromptIR}, \cite{cvpr2022AirNet}, \cite{NiPS2025_BioIR}.
Early attempts commonly relied on multi-task learning strategies~\cite{cvpr2022Uformer},\cite{mul_task_chen2024learning},\cite{DA-CLIP_mul_task}, which employed a shared encoder with multiple degradation-specific decoders. 
However, such architectures often suffer from parameter redundancy and overlook the intrinsic relationships between different degradation types. 
Recent studies therefore focus on unified architectures that learn degradation-aware or degradation-invariant representations. 
For example, AirNet~\cite{cvpr2022AirNet} distinguishes degradation types through contrastive learning in a shared latent space, while DPPD~\cite{DynamicPrompt} and MWFormer~\cite{TIP2024MWformer} introduce dynamic parameter generation or degradation-aware guidance mechanisms. 
Despite encouraging progress, All-in-One restoration still faces the challenge of negative task interference, since different degradations often require conflicting restoration operations (e.g., smoothing for denoising versus sharpening for deblurring). 
Designing a unified framework that can effectively discriminate and adapt to diverse degradations without sacrificing performance therefore remains an important research problem.

\subsection{Prompt-Based Image Restoration}

Prompt learning~\cite{Prompt_Learn_zhou2022conditional} has recently emerged as an effective paradigm for guiding restoration models with degradation-aware priors. Several representative approaches, such as PromptIR~\cite{PromptIR}, AdaIR~\cite{ICLR2025_AdaIR}, and DPPD~\cite{DynamicPrompt}, introduce learnable prompts to encode degradation-specific information and modulate network representations. PromptIR~\cite{PromptIR} employs lightweight prompts as plugin modules to condition the network on different degradation types, while AdaIR~\cite{ICLR2025_AdaIR} leverages adaptively decoupled spectral information to enable frequency-aware restoration. DPPD~\cite{DynamicPrompt} further improves unified restoration by learning discriminative degradation prototypes and distributing prompt information across the network. Despite their effectiveness, most prompt-based restoration frameworks adopt a decoder-centric conditioning strategy, where prompts are attached as auxiliary modules to decoder layers. Such prompts are typically shared globally across spatial locations and influence features through static or additive modulation. As a result, prompt information is often decoupled from the internal attention computation, limiting the ability to explicitly regulate attention responses or suppress degradation-specific noise under complex or mixed degradations.

In contrast, our approach adopts a different formulation. We generate hierarchical prompts and a global contextual representation directly from the degraded input through a lightweight degradation-aware network. These prompts are jointly applied to both encoder and decoder stages, enabling consistent degradation-aware modulation throughout the feature hierarchy. Moreover, instead of simple additive fusion, degradation information are incorporated into the attention computation itself, allowing more fine-grained and context-aware feature regulation for complex restoration scenarios.

\subsection{Gating in Neural Networks}
Gating mechanisms have become a fundamental strategy for regulating information flow in deep neural networks. Originally conceptualized to filter temporal dynamics and stabilize gradients in recurrent models\cite{GRU_dey2017gate}, gating has systematically permeated feed-forward architectures to ensure robust optimization\cite{SwiGLU}. Today, it is a cornerstone of representation learning, explicitly modulating feature interactions. In attention and state-space paradigms, query-dependent gating dynamically refines information propagation and softmax activations, solidifying its indispensable role in enhancing feature selectivity and training stability across diverse network topologies. In computer vision, gating is commonly realized through channel-wise or spatial-wise feature modulation, such as SE blocks and context gating mechanisms used in modern image restoration networks (e.g., GFN\cite{2018_GFN}, NAFNet\cite{NAFNet}). These approaches suppress irrelevant features and enhance task-relevant responses via learned masks, which is particularly effective for handling spatially variant degradations. Qiu et al~\cite{qiu2025gatedattentionlargelanguage} further shows that integrating gating directly into the internal dynamics of attention can improve model stability, non-linearity, and sparsity. Inspired by these insights, we propose a dual-gating paradigm that jointly exploits spatial and channel-wise modulation to explicitly regulate feature propagation for unified image restoration. This coordinated design selectively attenuates complex degradations, enabling robust image recovery under diverse corruption scenarios.

\section{Proposed Method}
\begin{figure*}[t]
    \centering
    \includegraphics[width=\textwidth]{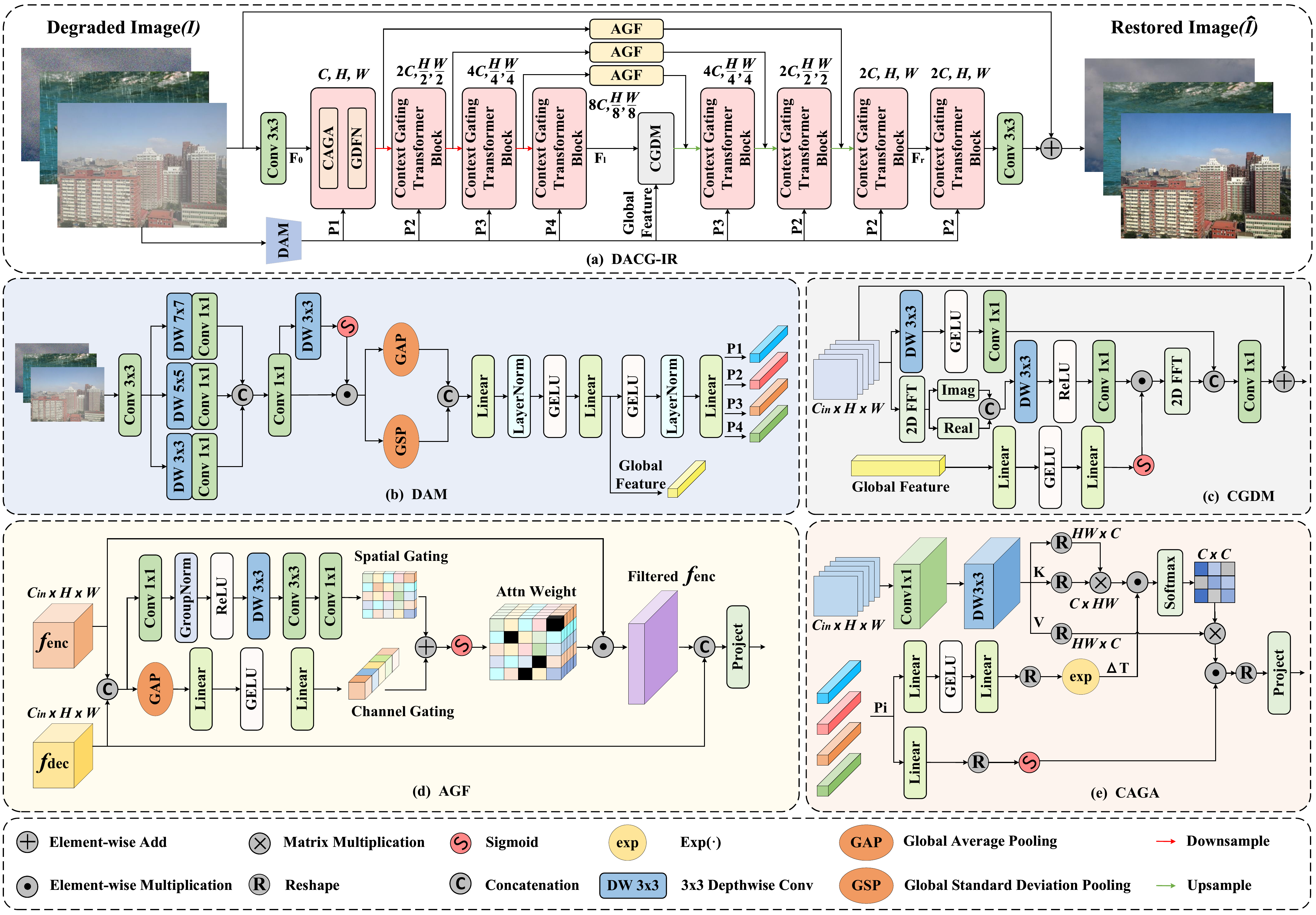}
    \caption{(a) Overview of the proposed DACG-IR architecture with four key components (b)–(e) for unified image restoration under diverse degradations. (b) Degradation-Aware Module (DAM) extracts multi-scale degradation statistics (e.g., mean and variance) to generate layer prompts and a global feature prompt, enabling adaptive modulation of restoration behaviors across different degradation levels. (c) Context Gated Dual-Domain Modulation (CGDM) performs joint spatial–frequency modulation by applying context-adaptive gating in the frequency domain. (d) Adaptive Gated Fusion (AGF) learns attention masks to dynamically reweight skip-connection features, emphasizing critical structures (e.g., edges and textures) and suppressing shallow noise propagation. (e) Context Adaptive Gated Attention (CAGA) introduces context-dependent temperature scaling and element-wise output gating into multi-head attention, allowing fine-grained and degradation-aware feature aggregation.}
    \label{fig:overall}
\end{figure*}

\subsection{Overall Pipeline}
Given a degraded input image $\mathbf{I}$, DACG-IR commences by extracting shallow features representations through a $3\times3$ convolution. Meanwhile, a degradation-aware context generation network produces multi-scale context prompts and a global degradation representation to encode degradation characteristics at different resolutions. Subsequently, the extracted features are processed by a four-level hierarchical encoder–decoder structure incorporating Context-Gating Transformer blocks~\cite{cvpr_2022_Restormer}. During encoding, the spatial resolution is progressively reduced while the channel dimension increases, and the corresponding prompts are injected at each stage to adaptively modulate feature transformation. At the bottleneck, by capturing long-range dependencies and frequency-aware information, a context-aware fusion module driven by the global degradation representation enhances latent features. The decoder then progressively upsamples the features to recover high-resolution representations. Instead of conventional skip connections, an Adaptive Gated Fusion (AGF) module selectively fuses encoder and decoder features to suppress degradation noise while preserving structural details, followed by further refinement with prompt-guided Transformer blocks. Finally, several refinement blocks are applied at the highest resolution, and a convolution layer with residual learning generates the restored image. Overall, DACG-IR conducts degradation-aware and context-guided image restoration in an end-to-end framework, allowing robust adaptation to diverse degradation patterns.

\subsection{Degradation-Aware Module}

To infer degradation priors for unified image restoration, we design a \emph{Degradation-Aware Module (DAM)} that learns both a global degradation representation and layer context prompts in a data-driven manner. As illustrated in Fig.~\ref{fig:overall}(b), for a degraded input image $\mathbf{I}$, an initial feature map $\mathbf{F}_{\text{in}} \in \mathbb{R}^{H \times W \times C}$ is first obtained using a $3\times3$ convolution, where $H$, $W$, and $C$ represent the spatial resolution and channel dimension.

To capture degradation patterns under different receptive fields, $\mathbf{F}_{\text{in}}$ is processed by $S$ parallel multi-scale branches. 
In the $s$-th branch, a depth-wise convolution with kernel size $k_s = 2s + 3$ extracts scale-specific features, followed by a point-wise convolution for channel interaction:
\begin{equation}
\mathbf{F}_s =
\text{Conv}_{1\times1}^{(s)}
\!\left(
\text{DW}_{k_s\times k_s}^{(s)}(\mathbf{F}_{\text{in}})
\right),
\label{eq:dam_multiscale}
\end{equation}
where $\text{DW}_{k_s\times k_s}$ and $\text{Conv}_{1\times1}$ denote depth-wise and point-wise convolutions, respectively.

The multi-scale feature set $\{\mathbf{F}_s\}_{s=0}^{S-1}$ is first concatenated along the channel dimension and processed by a $1\times1$ convolution to yield the fused representation $\mathbf{F}_{\text{fuse}}$. Subsequently, an adaptive spatial gating mechanism is employed to suppress responses irrelevant to the degradation:
\begin{equation}
\mathbf{F}_{\text{gated}} =
\mathbf{F}_{\text{fuse}} \odot
\sigma\!\left(
\text{DW}_{3\times3}(\mathbf{F}_{\text{fuse}})
\right),
\label{eq:dam_gating}
\end{equation}
where $\odot$ represents element-wise multiplication and $\sigma(\cdot)$ denotes the sigmoid function.

Based on the gated features, global degradation characteristics are encoded via dual-statistic pooling. Specifically, the spatial mean and standard deviation are computed and concatenated to form a global descriptor:
\begin{equation}
\mathbf{z}_{\text{stat}} =
\operatorname{Concat}
\bigl(
\mu(\mathbf{F}_{\text{gated}}),\,
\sigma_{\mathrm{std}}(\mathbf{F}_{\text{gated}})
\bigr).
\label{eq:dam_statistics}
\end{equation}

The descriptor is then processed by a multi-layer perceptron to obtain a compact global degradation feature that captures overall degradation intensity and variability. Conditioned on this feature, DAM generates hierarchical layer prompts $\{\mathbf{P}_i\}$ via successive linear projections with normalization and non-linearity. These prompts are aligned with the channel dimensions of different stages in the restoration network ($C$, $2C$, $4C$, and $8C$), providing multi-scale degradation-aware guidance for adaptive feature modulation.

\subsection{Context Gated Dual-Domain Modulation (CGDM)}

Deep latent features encode compact yet semantically rich representations, making them suitable for modeling global image degradations. However, addressing global artifacts (e.g., color shifts, haze, and illumination distortions) solely in the spatial domain is inefficient, since convolutional operators have limited receptive fields. To address this limitation, we introduce the \emph{Context Gated Dual-Domain Modulation} (CGDM) module, deployed at the latent bottleneck (Fig.~\ref{fig:overall}(c)). This design leverages high-level semantic representations while reducing computational cost for frequency-domain processing.

CGDM adopts a dual-branch architecture that jointly exploits spatial and frequency-domain characteristics. Given a latent feature map $\mathbf{F}_l \in \mathbb{R}^{8C \times \frac{H}{8} \times \frac{W}{8}}$, the spatial branch preserves local structural details such as edges and textures via depth-wise convolution followed by point-wise projection:
\begin{equation}
    \mathbf{Y}_{\mathrm{spa}} =
    \text{Conv}_{1\times1}
    \!\left(
    \delta\!\left(
    \text{DW}_{3\times3}(\mathbf{F}_l)
    \right)
    \right),
\end{equation}
where $\delta(\cdot)$ denotes GELU and $\text{DW}_{3\times3}$ represents a depth-wise convolution.

The frequency branch simultaneously models global degradation characteristics by mapping $\mathbf{F}_l$ to the spectral domain with a 2D Fourier transform. The real and imaginary components are concatenated and processed through a learnable mixing function. A degradation-aware global context embedding $P_{\mathrm{global}}$ generates a channel-wise spectral modulation mask to adaptively adjust frequency responses:
\begin{equation}
\begin{aligned}
    \mathbf{M}_{\mathrm{freq}} &= 
    \sigma\!\left(\mathcal{W}_{\mathrm{gate}} (P_{\mathrm{global}})\right),
    \quad \mathbf{M}_{\mathrm{freq}} \in \mathbb{R}^{2C \times 1 \times 1}, \\
    \mathbf{Y}_{\mathrm{freq}} &= 
    \mathcal{F}^{-1}\!\left(
    \mathcal{W}_{\mathrm{f}}
    \left(
    [\mathrm{Re}(\mathcal{F}(\mathbf{F}_l)), 
     \mathrm{Im}(\mathcal{F}(\mathbf{F}_l))]
    \right)
    \odot \mathbf{M}_{\mathrm{freq}}
    \right),
\end{aligned}
\end{equation}
where $\sigma(\cdot)$ denotes the sigmoid function, $\mathcal{F}(\cdot)$ and $\mathcal{F}^{-1}(\cdot)$ represent the Fourier and inverse Fourier transforms, $[\cdot]$ denotes channel concatenation, and $\mathcal{W}{\mathrm{gate}}(\cdot)$ and $\mathcal{W}{\mathrm{f}}(\cdot)$ are learnable projections.

Finally, spatial and frequency representations are fused to simultaneously capture local details and correct global degradations:
\begin{equation}
    \mathbf{Y}_{\mathrm{out}} =
    \mathbf{F}_l +
    \text{Conv}_{1\times1}
    \!\left(
    \mathrm{Concat}
    [\mathbf{Y}_{\mathrm{spa}}, \mathbf{Y}_{\mathrm{freq}}]
    \right),
\end{equation}
where $\text{Conv}_{1\times1}$ denotes a $1\times1$ convolution layer. This dual-domain design effectively balances local structure preservation and global degradation modeling for robust image restoration.

\subsection{Adaptive Gated Fusion}
Standard U-shaped image restoration uses skip connections\cite{PromptIR},\cite{NAFNet},\cite{cvpr_2022_Restormer}, to recover spatial details by transferring encoder features to the decoder. Naive fusion operations like concatenation or element-wise summation unselectively propagate encoder representations. Unfiltered transmission of encoder features containing degradation-related components disrupts decoder reconstruction and accumulates artifacts. These naive strategies ignore semantic discrepancies and non-uniform degradation distributions across spatial and channel dimensions. We propose an \emph{Adaptive Gated Fusion} (AGF) module (see Fig.~\ref{fig:overall}(d)) to rectify features by selectively preserving informative content and suppressing degradation-prone responses.

AGF is a plug-and-play component for U-shaped architectures. Given encoder features $\mathbf{F}_{\text{enc}} \in \mathbb{R}^{C \times H \times W}$ and decoder features $\mathbf{F}_{\text{dec}} \in \mathbb{R}^{C \times H \times W}$, the joint representation is:
\begin{equation}
    \mathbf{F}_{\text{cat}} = \text{Concat}(\mathbf{F}_{\text{enc}}, \mathbf{F}_{\text{dec}}) \in \mathbb{R}^{2C \times H \times W}.
\end{equation}
The spatial branch captures local degradation patterns by processing features through a lightweight sequence of operations, producing a spatial importance map that highlights structurally relevant regions. This design effectively exploits local contextual information while maintaining minimal computational overhead.
\begin{equation}
    \mathbf{S}' = \text{ReLU}(\text{GN}(\text{Conv}_{1\times1}(\mathbf{F}_{\text{cat}}))),
\end{equation}
\begin{equation}
    \mathbf{S} = \text{Conv}_{1\times1}(\text{Conv}_{3\times3}(\text{DW}_{3\times3}(\mathbf{S}')),
\end{equation}
where $\text{GN}$ denotes Group Normalization, $\text{Conv}{1\times1}$ represents $1\times1$ convolution layer, and $\text{DW}{3\times3}$ indicates a $3\times3$ depth-wise convolution. This structure expands the receptive field to identify spatially localized degradation. The channel gated branch models global reliability using a squeeze-and-excitation mechanism:
\begin{equation}
    \mathbf{C} = \text{Linear}(\text{ReLU}(\text{Linear}(\text{GAP}(\mathbf{F}_{\text{cat}})))),
\end{equation}
where $\text{GAP}$ is the global average pooling operation. These dual responses are integrated into a unified gated mask to modulate the encoder features:
\begin{equation}
    \mathbf{A} = \sigma(\mathbf{S} + \mathbf{C}),
\end{equation}
\begin{equation}
    \mathbf{F}_{\text{enc}}^{\prime} = \mathbf{F}_{\text{enc}} \odot \mathbf{A},
\end{equation}
where $\odot$ represents element-wise multiplication and $\sigma$ denotes the Sigmoid activation. The gated mask $\mathbf{A}$ is used to suppress regions dominated by degradations while maintaining key structural information.
The noise-filtered encoder features are ultimately fused with the original decoder representations to produce the final output:
\begin{equation}
    \mathbf{F}_{\text{out}} = \text{GELU}(\text{Conv}_{1\times1}(\text{Concat}(\mathbf{F}_{\text{enc}}^{\prime}, \mathbf{F}_{\text{dec}}))).
\end{equation}
The information flow through skip connections is dynamically adjusted based on input content to ensure robust feature integration across diverse degradation conditions.

\subsection{Context Adaptive Gated Attention}

Traditional Multi-Head Self-Attention (MHSA)~\cite{DFPIR},\cite{DynamicPrompt},\cite{cvpr_2022_Restormer},\cite{ICLR2025_AdaIR}, adopts a fixed scaling factor $\frac{1}{\sqrt{d_k}}$, implicitly assuming uniform feature reliability across layers and degradation levels. However, in image restoration, features are often contaminated by diverse degradations, making a static attention sharpness suboptimal. Severe degradation introduces noisy correlations that require smoother aggregation, whereas mild degradations benefit from sharper attention responses. To address this issue, we propose \emph{Context-Adaptive Gated Attention} (CAGA), shown in Fig.~\ref{fig:overall}(e), which utilizes the layer-wise degradation prior $P_i$ generated by DAM to regulate both the attention distribution and the output response.

\noindent\textbf{Context-Adaptive Temperature Scaling.}
Instead of using a fixed temperature, we introduce a context-adaptive temperature to control the entropy of attention distributions. Given query $\mathbf{Q}$ and key $\mathbf{K}$, $L_2$ normalization is first applied to stabilize the dot-product magnitude. A head-specific temperature $\tau_h$ is then predicted from the degradation-aware context:
\begin{equation}
    \tau_h = \exp\left(\theta_{\text{base}} + \mathcal{W}_{\tau}(P_i)\right),
\end{equation}
where $\theta_{\text{base}}$ is a learnable global bias and $\mathcal{W}_{\tau}$ maps the context prior to per-head modulation. The attention output of the $h$-th head is computed as
\begin{equation}
\text{Attention}(\mathbf{Q}_h,\mathbf{K}_h,\mathbf{V}_h)
= \text{softmax}\left(\frac{\mathbf{Q}_h \mathbf{K}_h^\top}{\tau_h}\right)\mathbf{V}_h .
\end{equation}
Adaptive temperature modulation enables the model to dynamically adjust attention sharpness, improving robustness under varying degradation levels.

\noindent\textbf{Gated Attention Output.}
Although temperature scaling controls attention formation, aggregated features may still contain degradation noise. Therefore, we introduce a context-aware gated mechanism on the attention output to suppress degradation-dominated responses. The gated attention is formulated as
\begin{equation}
    \mathbf{O}_{gated} = \mathbf{O}_{attn} \odot \sigma(\text{Linear}(P_i)),
\end{equation}
where $P_i$ denotes the layer-wise degradation prompt generated from the input image. This gated operation performs degradation-aware feature modulation by suppressing noisy responses while preserving restoration-relevant information.

\section{Experiments}
\subsection{Experiment Setup}

Following prior works~\cite{cvpr_2022_Restormer,PromptIR}, we adopt a four-level hierarchical encoder-decoder architecture. Two variants are implemented: DACG-IR and a lightweight version DACG-IR-S, which share a common topology but are configured with initial channel widths of 48 and 32, respectively. The encoder contains $[4,6,6,8]$ transformer blocks from shallow to deep stages with progressively increasing channels, while the decoder mirrors this configuration. A degradation-aware context module generates multi-scale prompts that are injected into each transformer block for adaptive feature modulation. In addition, adaptive gated fusion is applied to all skip connections, and a context-guided dual-domain module is introduced at the bottleneck to jointly model spatial and spectral representations. Before the reconstruction layer, a refinement stage consisting of four Transformer blocks is applied. The proposed network is trained end-to-end from scratch, utilizing the Adam optimizer with hyper-parameters $\beta_1=0.9$ and $\beta_2=0.999$. In accordance with~\cite{cvpr2025MoCE_IR}, we supervise the reconstruction process using an $\ell_1$ loss calculated in both the RGB and Fourier domains. We initialize the learning rate at $2\times10^{-4}$ and employ a cosine decay schedule, maintaining a batch size of 32 throughout the training phase. For the all-in-one restoration task, we utilize a crop size of $128\times128$, training the model for 100 and 120 epochs under the three- and five-degradation settings, respectively. For single-task, adverse-weather, and composite degradation scenarios, the crop size is expanded to $256\times256$ to ensure consistency with~\cite{PromptIR}. The entire pipeline is implemented in PyTorch and executed on NVIDIA Tesla A100 GPUs.

\noindent \textbf{Datasets.}
\textit{All-in-One setting:} Following the experimental setup in~\cite{PromptIR}, our unified restoration model is trained on a combination of datasets across multiple degradation types. For the three-task, we adopt SOTS (RESIDE)~\cite{SOTS} for dehazing and Rain100L~\cite{Rain100L} for deraining, Regarding denoising, noisy samples are synthesized by injecting Gaussian Noise with varying intensities ($\sigma = 15, 25, 50$) into the BSD400~\cite{BSD400} and WED~\cite{WED} datasets. Building upon this, the five-task setting further includes GoPro~\cite{GoPro} for deblurring and  LOL-v1~\cite{LOL_Retinex-Net} for low-light enhancement. All models are subsequently evaluated on the corresponding validation sets to ensure a fair comparison.

\textit{Single-Degradation setting:} For the single-task setting, an individual model is trained for each restoration task. The evaluation is conducted on Rain100L~\cite{Rain100L} and SPANet~\cite{SPAData} for deraining, SOTS-Outdoor and SOTS-Indoor~\cite{SOTS} for dehazing, BSD68~\cite{CBSD68}, Urban100~\cite{Urban100}, and Kodak24~\cite{Kodak24} for denoising, and GoPro~\cite{GoPro} for deblurring.

\textit{Multi-Weather setting:} We use All-Weather~\cite{All-in-One} to train and evaluate the adverse weather removal task, covering conditions such as rain + fog, raindrops, and snow.

\textit{composite degradations:} To assess the model on challenging compound degradations, experiments are conducted on the CDD11 dataset~\cite{CDD11}. 
\textbf{More dataset details are provided in the supplementary material.}
\begin{table*}[t]
\centering
\caption{Performance comparison between the proposed DACG-IR and state-of-the-art methods on \textbf{\textit{three}} degradation tasks.The best and second-best results are highlighted in bold and underline, respectively.}
\label{tab:three_degradations}
\resizebox{\textwidth}{!}{
\begin{tabular}{l l c c c c c c c}
\toprule
\multirow{2}{*}{\textbf{Type}} 
& \multirow{2}{*}{\textbf{Method}} 
& \multirow{2}{*}{\textbf{Dehazing}} 
& \multirow{2}{*}{\textbf{Deraining}} 
& \multicolumn{3}{c}{\textbf{Denoising (CBSD68\cite{CBSD68})}} 
& \multirow{2}{*}{\textbf{Average}} 
& \multirow{2}{*}{\textbf{Params (M)}} \\
\cmidrule(lr){5-7}
& & SOTS\cite{SOTS}& Rain100L\cite{Rain100L} & $\sigma=15$ & $\sigma=25$ & $\sigma=50$ & & \\
\midrule
\multirow{7}{*}{General}
& MPRNet\cite{MPRNet}~(CVPR'21)          & 28.00/0.958 & 33.86/0.958 & 33.27/0.920 & 30.76/0.871 & 27.29/0.761 & 30.63/0.894  & 15.74M \\
& Restormer\cite{cvpr_2022_Restormer}~(CVPR'22)       & 27.78/0.958 & 33.78/0.958 & 33.72/0.930 & 30.67/0.865 & 27.63/0.792 & 30.75/0.901  & 26.13M \\
& NAFNet\cite{NAFNet}~(ECCV'22)          & 24.11/0.928 & 33.64/0.956 & 33.03/0.918 & 30.47/0.865 & 27.12/0.754 & 29.67/0.844  & 17.11M \\
& DRSFormer\cite{DRSFormer}~(CVPR'23)        & 29.02/0.968 & 35.89/0.970 & 33.28/0.921 & 30.55/0.862 & 27.58/0.786 & 31.26/0.902  & 33.72M \\
& MambaIR\cite{MambaIR}~(ECCV'24)         & 29.57/0.970 & 35.42/0.969 & 33.88/0.931 & 30.95/0.874 & 27.74/0.793 & 31.51/0.907  & 26.78M \\
\midrule
\multirow{11}{*}{All-in-One}
& AirNet\cite{cvpr2022AirNet}~(CVPR'22)          & 27.94/0.962 & 34.90/0.967 & 33.92/0.932 & 31.26/0.888 & 28.00/0.797 & 31.20/0.910  & 8.93M \\
& NDR\cite{NDR}~(TIP'24)              & 28.64/0.962 & 35.42/0.969 & 34.01/0.932 & 31.36/0.887 & 28.10/0.798 & 31.51/0.910 & 28.40M \\
& InstructIR\cite{InstructIR}~(ECCV'24)     & 30.22/0.959 & 37.98/0.978 & 34.15/0.933 & 31.52/0.890 & 28.30/0.804 & 32.43/0.913  & 15.84M \\
& PromptIR\cite{PromptIR}~(NeurIPS'23)     & 30.58/0.974 & 36.37/0.972 & 33.98/0.933 & 31.31/0.888 & 28.06/0.799 & 32.06/0.913 & 35.59M \\
& MoCE-IR\cite{cvpr2025MoCE_IR}~(CVPR'25)         & 31.34/0.979 & 38.57/\underline{0.984} & 34.11/0.932 & 31.45/0.888 & 28.18/0.800 & 32.73/0.917 & 25.35M \\
& DFPIR\cite{DFPIR}~(CVPR'25)            & \textbf{31.87}/0.980 & 38.65/0.982 & 34.14/0.935 & 31.47/\underline{0.893} & 28.25/0.806 & 32.88/0.919 & 31.10M \\
& AdaIR\cite{ICLR2025_AdaIR}~(ICLR'25)           & 31.06/0.980 & 38.64/0.983 & 34.12/0.935 & 31.45/0.892 & 28.19/0.802 & 32.69/0.918 & 28.77M \\
& DA-RCOT\cite{DA-RCOT}~(TPAMI'25)       & 31.26/0.977 & 38.36/0.983 & 33.98/0.934 & 31.33/0.890 & 28.10/0.801 & 32.60/0.917  & 50.90M \\
& Perceive-IR\cite{tip2025Perceive_IR}~(TIP'25)      & 30.87/0.975 & 38.29/0.980 & 34.13/0.934 & 31.53/0.890 & \underline{28.31}/0.804 & 32.63/0.917 & 42.02M \\
& BioIR-L\cite{NiPS2025_BioIR}~(NeurIPS'25)      & 31.69/0.981 & 38.63/\underline{0.984} & 34.19/\underline{0.937} & 31.54/\textbf{0.895} & 28.28/\textbf{0.809} & 32.87/\textbf{0.921} & 15.85M \\
\midrule
& \textbf{DACG-IR-S (Ours)}   & \underline{31.82}/\underline{0.982} & \underline{39.07}/\textbf{0.986} & \underline{34.21}/0.933 & \underline{31.55}/0.890 & 28.29/0.802 & \underline{32.99}/0.919 & 13.85M \\
& \textbf{DACG-IR (Ours)}     & 31.60/\textbf{0.983} & \textbf{39.28}/\textbf{0.986} & \textbf{34.25}/\textbf{0.938} & \textbf{31.60}/0.892 & \textbf{28.35}/\underline{0.808} & \textbf{33.02}/\textbf{0.921} & 30.86M \\
\bottomrule
\end{tabular}}
\end{table*}
\begin{table*}[t]
\centering
\captionsetup{justification=centering}
\caption{Performance comparison between the proposed DACG-IR and state-of-the-art methods on \textbf{\textit{five}} degradation tasks.The best and second-best results are highlighted in bold and underline, respectively.}
\label{tab:five_degradations}
\resizebox{\textwidth}{!}{
\begin{tabular}{l l ccccccc}
\toprule
\multirow{2}{*}{\textbf{Type}} 
& \multirow{2}{*}{\textbf{Method}} 
& \textbf{Dehazing} 
& \textbf{Deraining} 
& \textbf{Denoising} 
& \textbf{Deblurring} 
& \textbf{Low-light}
& \multirow{2}{*}{\textbf{Average}} 
& \multirow{2}{*}{\textbf{Params}} \\
\cmidrule(lr){3-3}\cmidrule(lr){4-4}\cmidrule(lr){5-5}\cmidrule(lr){6-6}\cmidrule(lr){7-7}
&  
& SOTS\cite{SOTS} 
& Rain100L \cite{Rain100L}
& CBSD68\cite{CBSD68} ($\sigma=25$) 
& GoPro\cite{GoPro}
& LOL\cite{LOL_Retinex-Net} 
&  &  \\
\midrule
\multirow{8}{*}{General}
& Restormer\cite{cvpr_2022_Restormer}~(CVPR'22)           & 24.09/0.927 & 34.81/0.962 & 31.49/0.884 & 27.22/0.829 & 20.41/0.806 & 27.60/0.881 & 26.13M \\
& NAFNet\cite{NAFNet}~(ECCV'22)              & 25.23/0.939 & 35.56/0.967 & 31.02/0.883 & 26.53/0.808 & 20.49/0.809 & 27.76/0.881 & 17.11M \\
& DRSformer\cite{DRSFormer}~(CVPR'23)           & 24.66/0.931 & 33.45/0.953 & 30.97/0.881 & 25.56/0.780 & 21.77/0.821 & 27.28/0.873 & 33.72M \\
& Retinexformer\cite{Retinexformer_cai2023retinexformer}~(ICCV'23)       & 24.81/0.933 & 32.68/0.940 & 30.84/0.880 & 25.09/0.779 & 22.76/0.834 & 27.24/0.873 & 1.61M \\
& MambaIR\cite{MambaIR}~(ECCV'24)             & 25.81/0.944 & 36.55/0.971 & 31.41/0.884 & 28.61/0.875 & 22.49/0.832 & 28.97/0.901 & 26.78M \\
\midrule
\multirow{15}{*}{All-in-One}
& TransWeather\cite{cvpr2022transweather}~(CVPR'22)        & 21.32/0.885 & 29.43/0.905 & 29.00/0.841 & 25.12/0.757 & 21.21/0.792 & 25.22/0.836 & 37.93M \\
& AirNet\cite{cvpr2022AirNet}~(CVPR'22)              & 21.04/0.884 & 32.98/0.951 & 30.91/0.882 & 24.35/0.781 & 18.18/0.735 & 25.49/0.846 & 8.93M \\
& InstructIR\cite{InstructIR}~(ECCV'24)          & 27.10/0.956 & 36.84/0.973 & 31.40/0.887 & 29.40/0.886 & 23.00/0.836 & 29.55/0.907 & 15.84M \\
& PromptIR\cite{PromptIR}~(NeurIPS'23)         & 26.54/0.949 & 36.37/0.970 & 31.47/0.886 & 28.71/0.881 & 22.68/0.832 & 29.15/0.904 & 35.59M \\
& MoCE-IR\cite{cvpr2025MoCE_IR}~(CVPR'25)             & 30.48/0.974 & 38.04/0.982 & 31.34/0.887 & \textbf{30.05}/\textbf{0.899} & 23.00/0.852 & 30.58/0.919 & 25.35M \\
& DFPIR\cite{DFPIR}~(CVPR'25)                & \underline{31.64}/0.979 & 37.62/0.978 & 31.29/0.885 & 28.82/0.873 & \textbf{23.82}/0.843 & 30.64/0.913 & 31.10M \\
& AdaIR\cite{ICLR2025_AdaIR}~(ICLR'25)               & 30.53/0.978 & 38.02/0.981 & 31.35/0.889 & 28.12/0.858 & 23.00/0.845 & 30.20/0.910 & 28.77M \\
& DA-RCOT\cite{DA-RCOT}~(TPAMI'25)            & 30.96/0.975 & 37.87/0.980 & 31.23/0.888 & 28.68/0.872 & 23.25/0.836 & 30.40/0.911 & 50.90M \\
& Perceive-IR\cite{tip2025Perceive_IR}~(TIP'25)          & 28.19/0.964 & 37.25/0.977 & 31.44/0.887 & 29.46/0.886 & 22.88/0.833 & 29.84/0.909 & 42.02M \\
& BioIR-L\cite{NiPS2025_BioIR}~(NeurIPS'25)          & \textbf{31.77}/0.981 &38.75/\underline{0.985} & \underline{31.52}/\textbf{0.894} & \underline{29.61}/0.889 & 23.29/\underline{0.862} & \underline{30.99}/\textbf{0.922 }& 15.85M \\
\midrule
& \textbf{DACG-IR-S (Ours)}       & 31.40/\underline{0.980} & \underline{38.94}/\underline{0.985} & 31.51/0.889 & 29.11/0.881 & 22.84/0.858 & 30.76/0.919 & 13.85M \\
& \textbf{DACG-IR (Ours)}         & 31.63/\underline{0.981} & \textbf{39.24}/\textbf{0.986} & \textbf{31.57}/\underline{0.890} & 29.58/\underline{0.891} & \underline{23.35}/\textbf{0.864} & \textbf{31.07}/\textbf{0.922} & 30.86M \\
\bottomrule
\end{tabular}}
\end{table*}

\begin{table*}[t]
\centering
\caption{Quantitative comparison on CDD11~\cite{CDD11} under single, double and triple degradation settings. Each entry reports PSNR / SSIM.}
\label{tab:cdd11}
\setlength{\tabcolsep}{3pt}
\renewcommand{\arraystretch}{1.15}
\resizebox{\textwidth}{!}{
\begin{tabular}{c cccc ccccc cc c}
\toprule
\multirow{2}{*}{Method} 
& \multicolumn{4}{c}{CDD11-Single} 
& \multicolumn{5}{c}{CDD11-Double} 
& \multicolumn{2}{c}{CDD11-Triple} 
& \multirow{2}{*}{Average} \\
\cmidrule(lr){2-5} \cmidrule(lr){6-10} \cmidrule(lr){11-12}
& Rain(R) & Low(L) & Snow(S) & Haze(H) 
& H+R & L+S & L+H & H+S & L+R 
& L+H+R & L+H+S &  \\
\midrule
AirNet~\cite{cvpr2022AirNet}
& 26.55/0.891 & 24.83/0.778 & 26.79/0.919 & 24.21/0.951
& 22.21/0.868 & 23.29/0.723 & 23.23/0.779 & 23.29/0.901 & 22.82/0.710
& 21.80/0.708 & 22.24/0.725
& 23.75/0.814 \\

PromptIR~\cite{PromptIR} 
& 31.56/0.946 & 26.32/0.805 & 31.53/0.960 & 26.10/0.969
& 24.54/0.924 & 24.51/0.761 & 24.49/0.789 & 23.70/0.925 & 25.05/0.771
& 24.74/0.752 & 23.33/0.747
& 25.90/0.850 \\

WGWSNet~\cite{WGWSNet}
& 33.15/0.964 & 24.39/0.774 & 34.43/0.973 & 27.90/0.982
& 27.23/0.935 & 24.60/0.763 & 24.27/0.800 & 27.65/0.960 & 25.06/0.772
& 23.90/0.772 & 23.97/0.771
& 26.96/0.863 \\

WeatherDiff~\cite{WeatherDiff}
& 24.85/0.885 & 23.58/0.763 & 24.80/0.888 & 21.99/0.904
& 21.25/0.868 & 22.12/0.707 & 21.83/0.756 & 21.99/0.868 & 22.69/0.730
& 21.23/0.716 & 21.04/0.698
& 22.49/0.799 \\

OneRestore~\cite{CDD11}
& 33.40/0.964 & 26.48/0.826 & 34.31/0.973 & 32.52/0.990
& 29.99/0.957 & 25.19/0.789 & 25.79/0.822 & 30.21/0.964 & 25.58/0.799
& 24.78/0.788 & 24.90/0.791
& 28.47/0.878 \\

MoCE-IR-S~\cite{cvpr2025MoCE_IR}
& 34.31/\underline{0.970} & \underline{27.26}/0.824 & 35.91/\textbf{0.980} & 32.66/0.990
& 29.93/0.964 & \underline{26.04}/0.793 & \underline{26.24}/0.817 & 30.19/0.970 & \underline{26.25}/0.800
& \underline{25.41}/0.789 & \underline{25.39}/0.790
& 29.05/0.881 \\

BioIR-T~\cite{NiPS2025_BioIR}
& \textbf{34.48}/\underline{0.970} & 27.14/\textbf{0.834} & \textbf{36.26}/\underline{0.979} & \underline{34.77}/\underline{0.992}
& \textbf{31.64}/\underline{0.968} & 26.01/\textbf{0.804} & 26.11/\textbf{0.830} & \textbf{31.60}/\underline{0.972} & 26.17/\textbf{0.809}
& 25.22/\underline{0.799} & 25.34/\textbf{0.802}
& \underline{29.52}/\textbf{0.887} \\

\midrule
\textbf{DACG-IR-S (Ours)} 
& \underline{34.40}/\textbf{0.971} & \textbf{27.41}/\underline{0.827} & \underline{36.02}/\textbf{0.980} & \textbf{35.03/0.993}
& \underline{31.36}/\textbf{0.969} & \textbf{26.36}/\underline{0.800} & \textbf{26.89}/\underline{0.826} & \underline{31.57}/\textbf{0.974} & \textbf{26.50}/\underline{0.806}
& \textbf{25.94/0.799} & \textbf{25.91}/\underline{0.798}
& \textbf{29.76}/\underline{0.886} \\
\bottomrule
\end{tabular}
}
\end{table*}

\noindent \textbf{Evaluation Metrics:} To evaluate DACG-IR, we compare it with state-of-the-art methods under a unified experimental setup. All models are trained and tested using identical protocols to ensure fairness. Performance is measured by PSNR and SSIM on RGB channels (higher is better), along with visual comparisons for qualitative assessment.

\subsection{All-in-One Restoration Results}
\textit{Three Degradations:} DACG-IR under the three-task all-in-one setting, where a single model jointly handles dehazing, deraining, and denoising. As reported in Table~\ref{tab:three_degradations}, DACG-IR consistently outperforms representative general-purpose and all-in-one methods. Compared with BioIR~\cite{NiPS2025_BioIR}, DACG-IR improves the average PSNR by $0.15$~dB, and surpasses PromptIR~\cite{PromptIR}, Perceive-IR~\cite{tip2025Perceive_IR}, and DFPIR~\cite{DFPIR} by $0.96$~dB, $0.39$~dB, and $0.14$~dB, respectively, while using comparable or fewer parameters. In particular, DACG-IR shows notable gains on deraining and denoising, demonstrating its effectiveness in handling degradations with diverse statistical characteristics. Moreover, the lightweight variant DACG-IR-S achieves competitive performance with substantially fewer parameters, yielding higher average PSNR than several larger all-in-one models.

\textit{Five Degradations:}
We further assess DACG-IR in a more challenging five-task all-in-one setting, where a single model is trained on the merged datasets of all tasks and tested on their corresponding benchmarks. As reported in Table~\ref{tab:five_degradations}, DACG-IR delivers the best overall performance among all compared approaches. In terms of the average PSNR across the five tasks, it outperforms Perceive-IR~\cite{tip2025Perceive_IR} by $1.20$~dB and DA-RCOT~\cite{DA-RCOT} by $0.67$~dB, demonstrating strong robustness to diverse and complex degradations. In addition, the lightweight version, DACG-IR-S, achieves a favorable balance between restoration accuracy and computational efficiency. 

\begin{figure*}[t]
    \centering
    \includegraphics[width=\textwidth]{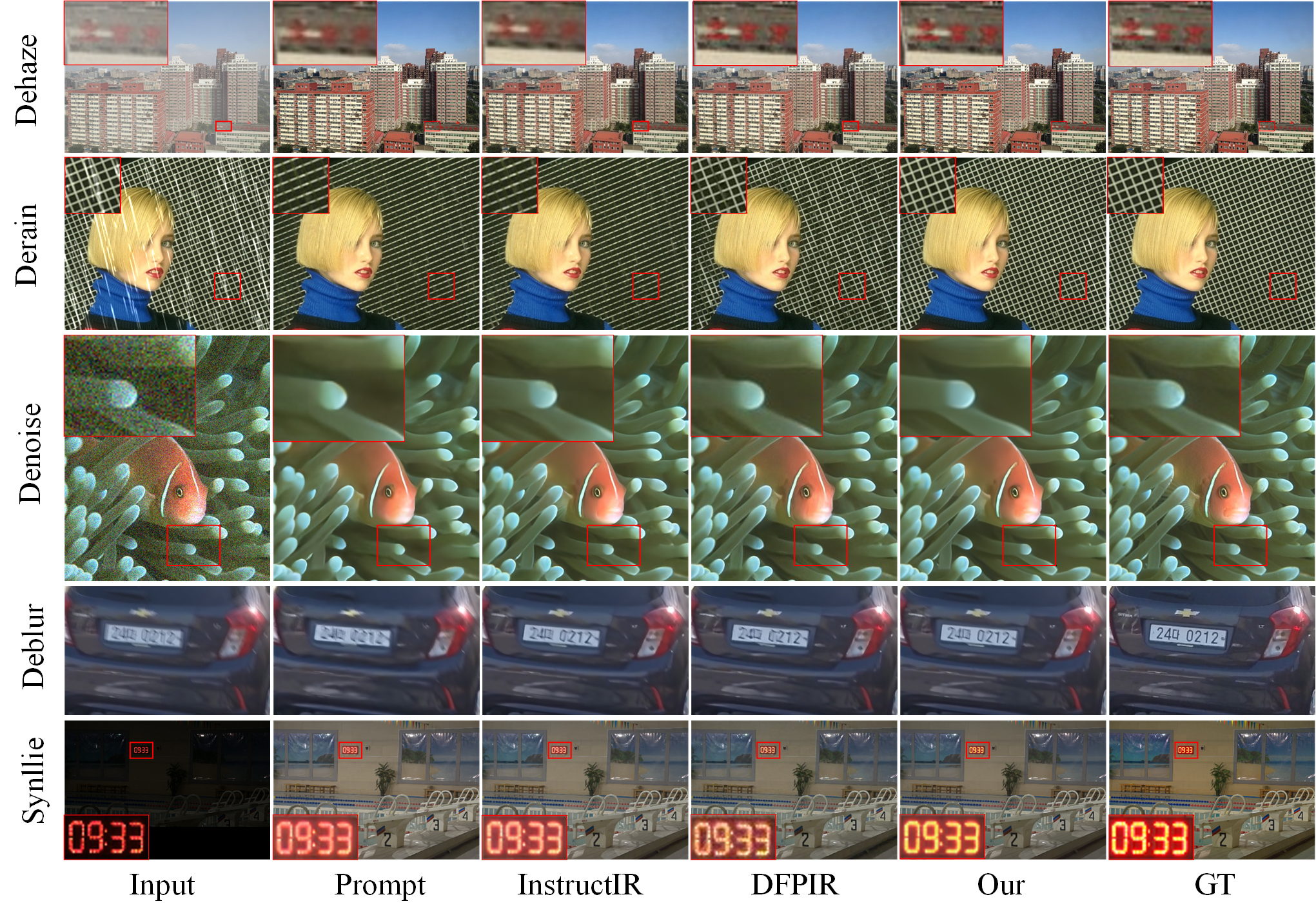}
    \caption{Visual comparison under the all-in-one setting with five degradation types. Compared with representative methods, including PromptIR\cite{PromptIR}, InstructIR\cite{InstructIR}, and DFPIR\cite{DFPIR}, DACG-IR produces visually superior restoration results with clearer structures and higher perceptual quality. Zoom in for better view.}
    \label{fig:five_Visual_Result}
\end{figure*}

Fig.~\ref{fig:five_Visual_Result} presents qualitative comparisons under five degradation types in the all-in-one setting. Compared with PromptIR~\cite{PromptIR}, InstructIR~\cite{InstructIR}, and DFPIR~\cite{DFPIR}, In both dehazing and deraining scenarios, our method yields reconstructions that exhibit superior visual fidelity to the ground truth, exhibiting more accurate structural recovery and fewer remaining artifacts, particularly in the deraining case where streak removal is more challenging. For motion deblurring and low-light enhancement, the proposed approach produces images with more natural color reproduction and improved visual clarity, leading to results that better preserve scene details and maintain higher consistency with the ground-truth images.

\begin{table*}[t]
    \centering
    \caption{The proposed DACG-IR and state-of-the-art methods on representative single-degradation restoration benchmarks.}
    \label{tab:signal_degradation_results}

    \subfloat{
        \begin{minipage}[t]{0.38\textwidth}
        \centering
        \textit{Dehazing} \\[0.2em]
        \begin{tabularx}{\textwidth}{lCC}
            \toprule
            Method & SOTS-outdoor & SOTS-indoor \\
            \midrule
            DehazeNet\cite{DehazeNet}   & 22.46/0.851 & 19.82/0.821 \\
            SGID-PPF\cite{SGID-PPF}    & 30.20/0.975 & 38.52/0.991 \\
            Dehazefomer\cite{Dehazefomer} & 31.45/0.978 & 38.46/0.994 \\
            \midrule
            Restormer\cite{cvpr_2022_Restormer}   & 30.87/0.969 & 38.88/0.991 \\
            NAFNet\cite{NAFNet}      & 30.98/0.970 & 38.97/0.992 \\
            FSNet\cite{FSNet-TPAMI}       & 31.11/0.971 & 40.47/\underline{0.996} \\
            \midrule
            PromptIR\cite{PromptIR}    & 31.31/0.973 & -- \\
            BioIR\cite{NiPS2025_BioIR}       & 31.69/\underline{0.981} & \underline{40.50}/\textbf{0.997} \\
            DFPIR\cite{DFPIR}       & \underline{32.00}/\underline{0.981} & -- \\
            \textbf{DACG-IR} & \textbf{32.21/0.983} & \textbf{40.52/}\underline{0.996} \\
            \bottomrule
        \end{tabularx}
        \end{minipage}
    }
    \hspace{0.8em}
    \subfloat{
        \begin{minipage}[t]{0.30\textwidth}
        \centering
        \textit{Deraining} \\[0.2em]
        \begin{tabularx}{\textwidth}{lCC}
            \toprule
            Method & Rain100L & SPA-Data \\
            \midrule
            MSPFN\cite{MSPFN}      & 32.40/0.933 & 43.43/0.984 \\
            FSNet\cite{FSNet-TPAMI}      & 36.56/0.974 & 41.02/0.980 \\
            DRSformer\cite{DRSFormer}  & 38.14/0.983 & \textbf{48.54}/\underline{0.992} \\
            \midrule
            AirNet\cite{cvpr2022AirNet}     & 34.90/0.977 & 39.67/0.969 \\
            Restormer\cite{cvpr_2022_Restormer}  & 36.74/0.978 & 41.39/0.981 \\
            DA-CLIP\cite{DA-CLIP_mul_task}    & 37.02/0.978 & 42.84/0.988 \\
            \midrule
            PromptIR\cite{PromptIR}   & 37.23/0.980 & 41.17/0.979 \\
            BioIR\cite{NiPS2025_BioIR}      & 38.63/\underline{0.984} & \underline{49.39}/\textbf{0.993} \\
            DFPIR\cite{DFPIR}      & \textbf{39.08}/\underline{0.984} & -- \\
            \textbf{DACG-IR} & \underline{39.06}/\textbf{0.985} & 48.36/\underline{0.992} \\
            \bottomrule
        \end{tabularx}
        \end{minipage}
    }
    \hspace{0.8em}
    \subfloat{
        \begin{minipage}[t]{0.22\textwidth}
        \centering
        \textit{Deblurring} \\[0.2em]
        \begin{tabularx}{\textwidth}{lC}
            \toprule
            Method & GoPro \\
            \midrule
            DeblurGAN\cite{DeblurGAN}   & 28.70/0.858 \\
            StripFormer\cite{Transformer_tsai2022stripformer} & \underline{33.08}/\underline{0.962} \\
            HI-Diff\cite{HI-Diff}     & \textbf{33.33/0.964} \\
            \midrule
            MPRNet\cite{MPRNet}      & 32.66/0.959 \\
            Restormer\cite{cvpr_2022_Restormer}   & 32.92/0.961 \\
            NAFNet\cite{NAFNet}      & 33.03/0.961 \\
            \midrule
            PromptIR\cite{PromptIR}    & 32.41/0.956 \\
            AirNet\cite{cvpr2022AirNet}      & 31.64/0.945 \\
            Perceive-IR\cite{tip2025Perceive_IR} & 32.83/0.960 \\
            \textbf{DACG-IR} & 33.06/0.959 \\
            \bottomrule
        \end{tabularx}
        \end{minipage}
    }
\end{table*}
\begin{table}[t]
\centering
\caption{Performance comparison on Gaussian image denoising. DACG-IR is compared with existing approaches. }
\label{tab:denoising_result}
\setlength{\tabcolsep}{2.0pt}
\renewcommand{\arraystretch}{1.05}
\footnotesize
\resizebox{\columnwidth}{!}{
\begin{tabular}{l|ccc|ccc|ccc}
\toprule
\multirow{2}{*}{Method} 
& \multicolumn{3}{c|}{Urban100} 
& \multicolumn{3}{c|}{CBSD68} 
& \multicolumn{3}{c}{Kodak24} \\
\cmidrule(lr){2-4} \cmidrule(lr){5-7} \cmidrule(lr){8-10}
& $\sigma$=15 & $\sigma$=25 & $\sigma$=50 
& $\sigma$=15 & $\sigma$=25 & $\sigma$=50 
& $\sigma$=15 & $\sigma$=25 & $\sigma$=50 \\
\midrule
Restormer~\cite{cvpr_2022_Restormer}    
& 33.72 & 31.26 & 28.03 
& 34.03 & 31.49 & 28.11 
& 34.78 & 32.37 & 29.08 \\

TransWeather~\cite{cvpr2022transweather}  
& 29.64 & 27.97 & 26.08 
& 31.16 & 29.00 & 26.08 
& 31.67 & 29.64 & 26.74 \\

InstructIR-5D~\cite{InstructIR} 
& 33.77 & 31.40 & 28.13 
& 34.00 & 31.40 & 28.15 
& 34.70 & 32.26 & 29.16 \\

PromptIR~\cite{PromptIR}     
& 34.77 & 32.49 & 29.39 
& 34.34 & 31.71 & 28.49 
& -- & -- & -- \\

MoCE-IR~\cite{cvpr2025MoCE_IR}      
& 34.01 & 31.59 & 28.20 
& 34.00 & 31.34 & 28.07 
& 34.87 & 32.38 & 29.20 \\

AdaIR~\cite{ICLR2025_AdaIR}        
& 34.10 & 31.68 & 28.29 
& 34.01 & 31.35 & 28.06 
& \underline{34.89} & 32.38 & \underline{29.21} \\

Perceive-IR~\cite{tip2025Perceive_IR}  
& \underline{34.86} & \underline{32.55} & \underline{29.42} 
& \textbf{34.38} & \underline{31.74} & \textbf{28.53} 
& 34.84 & \underline{32.50} & 29.16 \\

\textbf{DACG-IR} 
& \textbf{34.95} & \textbf{32.62} & \textbf{29.59} 
& \underline{34.35} & \textbf{31.75} & \underline{28.50}
& \textbf{35.33} & \textbf{32.91} & \textbf{29.86} \\
\bottomrule
\end{tabular}}
\end{table}

\subsection{Single Degradation Restoration Results}
To assess performance in single-degradation scenarios, we train task-specific models for each restoration task. Quantitative comparisons are reported in Table~\ref{tab:signal_degradation_results}. Overall, DACG-IR achieves the best or highly competitive performance across most tasks. For image dehazing, DACG-IR surpasses the task-specific Dehazeformer~\cite{Dehazefomer} by $0.76$~dB and $2.06$~dB on SOTS-outdoor and SOTS-indoor, respectively, and also outperforms several general restoration models such as FSNet~\cite{FSNet-TPAMI}. For deraining and motion deblurring, DACG-IR remains competitive with both general and all-in-one approaches, while showing a small gap compared with highly specialized task-specific models (e.g., DRSformer~\cite{DRSFormer} and HI-Diff~\cite{HI-Diff}).
Additional results are provided in the supplementary material.
\subsection{Multi-weather restoration Results}
We further evaluate DACG-IR on image restoration under three adverse weather degradations, including snow, rain streaks, and raindrops. Table~\ref{tab:Mul-Weather_comparison} presents the quantitative results on three widely adopted benchmarks. DACG-IR delivers the best overall performance across these datasets, outperforming both general-purpose restoration models and recent all-in-one approaches. On Snow100K-L~\cite{Snow100K_DesnowNet}, the SSIM achieved by DACG-IR is slightly lower than that of GridFormer~\cite{GridFormer}. Nevertheless, on Outdoor-Rain~\cite{Outdoor-Rain} and RainDrop~\cite{RainDrop}, DACG-IR achieves superior average performance, surpassing all existing state-of-the-art methods with the highest recorded PSNR and SSIM scores, DACG-IR exceeds the second-best method M2Restore~\cite{tip2025m2restore} by $0.46$~dB in PSNR and more than $0.002$ in SSIM, demonstrating its strong ability to handle diverse weather-related degradations.
\subsection{Composite Degradationsn Restoration Results}
DACG-IR is further evaluated on the CDD11~\cite{CDD11} composite degradation benchmark to assess its robustness and generalization. This setting involves eleven categories synthesized from concurrent distortions including low-light, haze, rain, and snow. As evidenced in Table~\ref{tab:cdd11}, DACG-IR-S yields the highest average PSNR, exceeding OneRestore~\cite{CDD11} by $1.29$~dB and outstripping the recent MoCE-IR-S~\cite{cvpr2025MoCE_IR} and BioIR-T~\cite{NiPS2025_BioIR} by $0.71$~dB and $0.24$~dB. These results confirm that by leveraging degradation-aware priors and isolating specific features in the latent space, DACG-IR-S effectively mitigates inter-feature interference and enhances restoration fidelity.

\begin{table*}[t]
\centering
\caption{Performance metrics on the All-Weather image restoration benchmark. Results are reported in PSNR (dB) and SSIM, with bold and underlined values denoting the top-performing and runner-up methods, respectively.}
\label{tab:Mul-Weather_comparison}
\setlength{\tabcolsep}{3.2pt}
\renewcommand{\arraystretch}{1.1}
\footnotesize
\begin{tabular}{c|l|cc|cc|cc|cc|c}
\hline
Type & Method  
& \multicolumn{2}{c|}{Outdoor-Rain~\cite{Outdoor-Rain}} 
& \multicolumn{2}{c|}{RainDrop\cite{RainDrop}} 
& \multicolumn{2}{c|}{Snow100K-L\cite{Snow100K_DesnowNet}} 
& \multicolumn{2}{c|}{Average} 
& Params \\
\hline
\multirow{6}{*}{General}
& SwinIR\cite{SwinIR_transformer} (ICCV'21)         & 23.23 & 0.869 & 30.82 & 0.904 & 28.18 & 0.880 & 27.41 & 0.884 & 1M  \\
& NAFNet\cite{NAFNet} (ECCV'22)         & 23.21 & 0.840 & 28.90 & 0.890 & 27.68 & 0.847 & 26.60 & 0.859 & 17M \\
& Uformer\cite{cvpr2022Uformer} (CVPR'22)       & 25.40 & 0.889 & 27.38 & 0.919 & 26.60 & 0.887 & 26.46 & 0.898 & 5M  \\
& Restormer\cite{cvpr_2022_Restormer} (CVPR'22)     & 27.24 & 0.921 & 29.29 & 0.937 & 27.76 & 0.907 & 28.10 & 0.922 & 26M \\
\hline
\multirow{10}{*}{All-in-One}
& All-in-One\cite{All-in-One} (CVPR'20)    & 24.71 & 0.898 & 31.12 & 0.927 & 28.33 & 0.882 & 28.05 & 0.902 & 44M \\
& TransWeather\cite{cvpr2022transweather} (CVPR'22)  & 28.83 & 0.900 & 30.17 & 0.916 & 29.31 & 0.888 & 29.44 & 0.901 & 38M \\
& WeatherDiff\cite{WeatherDiff} (TPAMI'23) & 29.72 & 0.922 & 29.66 & 0.923 & 29.58 & 0.894 & 29.65 & 0.913 & 83M \\
& WGWSNet\cite{WGWSNet} (CVPR'23)       & 25.31 & 0.901 & 31.31 & 0.932 & 29.71 & 0.894 & 28.78 & 0.909 & 6M  \\
& MWFormer\cite{TIP2024MWformer} (TIP'24)       & 29.07 & 0.901 & 31.09 & 0.922 & 30.05 & 0.899 & 30.07 & 0.907 & 170M \\
& GridFormer\cite{GridFormer} (IJCV'24)    & 30.48 & 0.931 & 31.02 & 0.930 & 30.78 & \textbf{0.917} & 30.76 & 0.926 & 30M \\
& M2Restore\cite{tip2025m2restore} (TIP'25)      & \underline{32.01} & \underline{0.955} & 31.73 & 0.943 & \underline{31.20} & 0.911 & 31.65 & \underline{0.936} & 248M \\
& \textbf{DACG-IR-S (Ours)} & 31.99 & 0.952 & \underline{31.96} & \underline{0.947} & 31.16 & 0.906 & \underline{31.70} & \underline{0.936} & 14M \\
& \textbf{DACG-IR (Ours)} & \textbf{32.58} & \textbf{0.957} & \textbf{32.22} & \textbf{0.950} & \textbf{31.54} & \underline{0.912} & \textbf{32.11} & \textbf{0.938} & 31M \\
\hline
\end{tabular}
\end{table*}
\begin{figure}[t]
\centering
\includegraphics[width=\columnwidth]{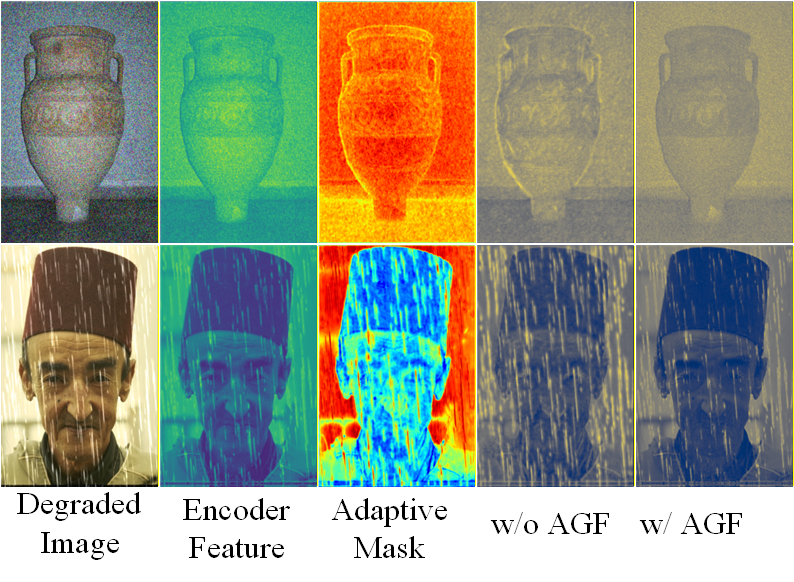}
\caption{Visualization of the Adaptive Gated Fusion (AGF) module. From left to right: degraded input, encoder features, learned adaptive mask, results without AGF, and results with AGF. The proposed AGF selectively suppresses degradation-related responses while enhancing structural and semantic details, leading to more faithful restorations.}
\label{fig:AGF_Visual_Result}
\end{figure}
\begin{figure}[t]
\centering
\includegraphics[width=\columnwidth]{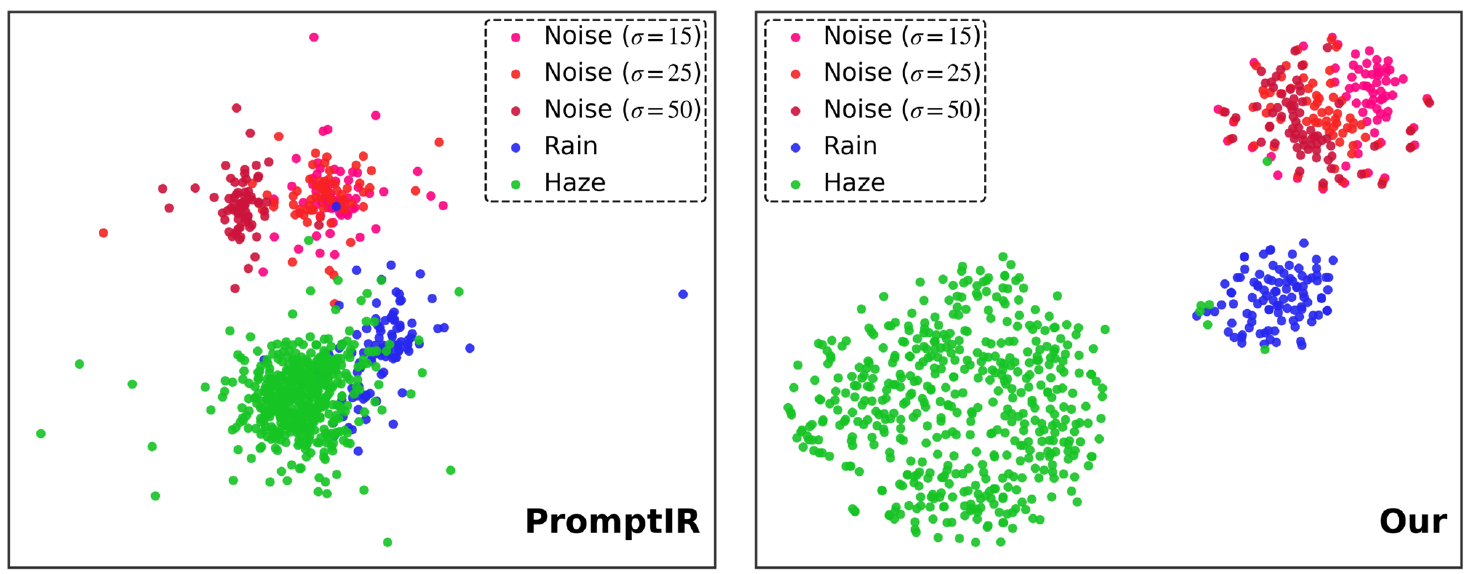}
\caption{The t-SNE visualization of intermediate features generated by PromptIR\cite{PromptIR} and our DACG-IR on the CBSD68\cite{CBSD68}, Rain100L\cite{Rain100L}, and SOTS\cite{SOTS} datasets. In the three-task setting, DACG-IR exhibits more distinct and cohesive task-specific clusters than PromptIR. This demonstrates that our method excels at extracting highly discriminative and degradation-aware representations for effective unified image restoration.}
\label{fig:t-SNE}
\end{figure}

\subsection{Ablation Study}
\subsubsection{Effect of Adaptive Gated Fusion (AGF)}
Designed as a plug-and-play module, the adaptive gated fusion (AGF) mechanism is employed to replace the original skip connections in existing U-shaped image restoration architectures, enabling an isolated evaluation of its effectiveness while keeping all other experimental settings unchanged. Extensive experiments are conducted across three representative degradation tasks, with quantitative results summarized in Table~\ref{tab:AGF_quantitative_comparison}. Initially, we integrate AGF into Restormer~\cite{cvpr_2022_Restormer} by substituting its skip connections. This modification yields an average improvement of 1.52~dB over the baseline, with especially significant gains of 2.47~dB and 4.00~dB on the dehazing and deraining tasks, respectively. These improvements can be primarily attributed to the spatial and channel gating operations in AGF, which adaptively suppress degradation noise in the encoder features while retaining essential structural details. To further demonstrate the versatility of AGF, we incorporate it into NAFNet~\cite{NAFNet} and PromptIR~\cite{PromptIR}. For NAFNet, the addition of AGF results in an average PSNR increase of 1.18~dB, including a remarkable 4.93~dB gain on the dehazing task. Similarly, integrating AGF into PromptIR enhances the average PSNR by 0.53~dB, with consistent improvements observed across all evaluated datasets. Figure~\ref{fig:AGF_Visual_Result} provides visual comparisons under challenging rainy and noisy conditions. While the encoder features in the baseline models still contain evident degradation noise, AGF effectively filters out these artifacts without compromising structural integrity. As highlighted in the fourth and fifth columns, directly passing noisy encoder features to the decoder substantially deteriorates restoration quality. In contrast, AGF enables the decoder to reconstruct cleaner and more detailed structures, demonstrating its efficacy in both noise suppression and feature preservation.

\subsubsection{Context Gated Dual-Domain Modulatio(CGDM)}
By transforming deep encoder features into the frequency domain from the spatial domain, CGDM mitigates the limited capability of CNNs and Transformers in exploiting frequency-domain information. Several degradations are more distinguishable in the frequency domain: for instance, periodic rain streaks produce distinct frequency peaks, while haze mainly occupies low-frequency components, making such global artifacts difficult to remove using spatial processing alone. Therefore, CGDM leverages the global degradation prompt generated by the Degradation-Aware Module (DAM) to perform context-guided frequency modulation on deep features.CGDM is inserted only after the deepest encoder stage, where features have a global receptive field and better align with the degradation-aware context embedding, allowing CGDM to act as a sample-adaptive global frequency filter. Moreover, operating at this low spatial resolution reduces computational cost and parameter overhead.As shown in Table~\ref{tab:ablation_result}, CGDM achieves 32.22~dB PSNR and 0.912 SSIM across three degradation tasks. When combined with Adaptive Gated Fusion (AGF), the performance further improves to 32.48~dB, and integrating Context Adaptive Gated Attention (CAGA) yields the best result of 32.61~dB. The t-SNE visualization in Fig.~\ref{fig:t-SNE} also shows more compact and well-separated task-wise feature distributions, indicating enhanced degradation-aware representations. These results demonstrate that CGDM effectively exploits frequency-domain information and complements spatial processing to suppress global degradations and improve restoration quality.
\begin{table*}[t]
\centering
\caption{Ablation study of AGF integrated into representative U-shaped-based image restoration models. Quantitative comparisons are conducted on dehazing, deraining, and denoising tasks. FLOPs are computed with an input resolution of 224×224.}
\label{tab:AGF_quantitative_comparison}
\setlength{\tabcolsep}{4pt}
\renewcommand{\arraystretch}{1.1}
\footnotesize
\begin{tabular}{lccccccccc}
\toprule
\multirow{2}{*}{Method} 
& \multirow{2}{*}{Params} 
& \multirow{2}{*}{FLOPs} 
& \multicolumn{1}{c}{Dehazing} 
& \multicolumn{1}{c}{Deraining} 
& \multicolumn{3}{c}{Denoising (CBSD68\cite{CBSD68})}  
& \multirow{2}{*}{Average} \\

\cmidrule(lr){4-4} \cmidrule(lr){5-5} \cmidrule(lr){6-8}

 &  &  
 & SOTS\cite{SOTS} 
 & Rain100L\cite{Rain100L} 
 & $\sigma$=15 
 & $\sigma$=25 
 & $\sigma$=50 
 &  \\
\midrule

Restormer\cite{cvpr_2022_Restormer} 
& 26.127M & 121.86G 
& 27.78/0.958 & 33.78/0.958 
& 33.72/0.930 & 30.67/0.865 & 27.63/0.792 
& 30.75/0.901 \\

\textbf{Restormer + AGF} 
& \textbf{26.359M} & \textbf{122.88G} 
& \textbf{30.25/0.977} & \textbf{37.78/0.980} 
& \textbf{34.04/0.931} & \textbf{31.38/0.886} & \textbf{28.09/0.795} 
& \textbf{32.31/0.914} \\
\midrule

NAFNet\cite{NAFNet} 
& 17.112M & 12.228G 
& 24.11/0.928 & 33.64/0.956 
& 33.03/0.918 & 30.47/0.865 & 27.12/0.754 
& 29.67/0.844 \\

\textbf{NAFNet + AGF} 
& \textbf{17.686M} & \textbf{13.304G} 
& \textbf{29.04/0.969} & \textbf{33.71/0.956} 
& \textbf{33.32/0.920} & \textbf{30.73/0.872} & \textbf{27.43/0.772} 
& \textbf{30.85/0.898} \\
\midrule
PromptIR\cite{PromptIR} 
& 35.592M & 135.17G 
& 30.58/0.974 & 36.37/0.972 
& 33.98/0.933 & 31.31/0.888 & 28.06/0.799 
& 32.06/0.913 \\

\textbf{PromptIR + AGF} 
& \textbf{35.862M} & \textbf{137.22G} 
& \textbf{31.09/0.979} & \textbf{38.12/0.981} 
& \textbf{34.11/0.928} & \textbf{31.39/0.890} & \textbf{28.22/0.801} 
& \textbf{32.59/0.916} \\
\bottomrule
\end{tabular}
\end{table*}
\begin{table}[t]
\centering
\caption{Ablation studies are conducted across three restoration tasks to investigate the impact of various module configurations. The quantitative results on full-resolution RGB images, reported in average PSNR (dB, ↑) and SSIM (↑), confirm that our full model configuration yields the optimal performance.}
\label{tab:ablation_result}
\setlength{\tabcolsep}{6pt}
\renewcommand{\arraystretch}{1.1}
\footnotesize
\begin{tabular}{c c c c c c}
\toprule
Method & AGF & CGDM & CAGA & PSNR & SSIM \\
\midrule
(a) & $\checkmark$ &  &  & 32.31 & 0.914 \\
(b) &  & $\checkmark$ &  & 32.22 & 0.912 \\
(c) &  &  & $\checkmark$ & 32.35 & 0.914 \\
(d) & $\checkmark$ & $\checkmark$ &  & 32.48 & 0.915 \\
(e) &  & $\checkmark$ & $\checkmark$ & 32.61 & 0.916 \\
(f) & $\checkmark$ &  & $\checkmark$ & 32.79 & 0.917 \\
\midrule
DACG-IR & $\checkmark$ & $\checkmark$ & $\checkmark$ & \textbf{33.02} & \textbf{0.921} \\
\bottomrule
\end{tabular}
\end{table}
\begin{table}[t]
\centering
\caption{Ablation study of Adaptive Temperature ($T$) and Gated Output ($G$) within CAGA across three restoration benchmarks. The quantitative results on full-resolution RGB images, reported in average PSNR (dB, ↑) and SSIM (↑).}
\label{tab:ablation_CAGA}
\setlength{\tabcolsep}{4pt}
\renewcommand{\arraystretch}{1.1}
\begin{tabular}{lccccc}
\toprule
Method & Adaptive Temp ($T$) & Gated Output ($G$) & PSNR & SSIM \\
\midrule
Baseline     &                    &                    & 30.75 & 0.901 \\
(b)     & \checkmark         &                    & 31.87 & 0.906 \\
(c)     &                    & \checkmark         & 32.06 & 0.911 \\
CAGA & \checkmark         & \checkmark         & \textbf{32.35} & \textbf{0.914} \\
\bottomrule
\end{tabular}
\end{table}
\subsubsection{Context Adaptive Gated Attention(CAGA)}
Context Adaptive Gated Attention (CAGA) is one of the key components in our framework.
We evaluate two core designs in the attention module, including the adaptive temperature coefficient ($T$) and the gated attention output ($G$). The ablation results are reported in Table \ref{tab:ablation_result} and Table \ref{tab:ablation_CAGA}. Compared with standard self-attention (SA), which adopts a fixed temperature and fully passes attention responses, CAGA dynamically modulates the temperature according to degradation prompts generated by DAM. When the input contains severe noise, a lower temperature is applied to smooth the attention distribution and reduce responses to unreliable regions. When the degradation is mild, a higher temperature is used to sharpen attention and preserve fine details. As shown in Table \ref{tab:ablation_CAGA}, the PSNR rises from the baseline SA's 30.75~dB to 31.87~dB with the sole addition of adaptive temperature modulation, validating the effectiveness of adaptive temperature control. When the gated output mechanism is applied, the performance gain reaches 1.31~dB over SA, which indicates that standard attention outputs contain redundant noise that degrades restoration quality. The gated mechanism suppresses noisy responses and leads to clearer reconstruction results. When adaptive temperature modulation and gated output are jointly used, the average PSNR further improves to 32.35~dB, which is higher than using either component alone. This result shows that the two mechanisms work together to enhance attention reliability. Furthermore, the ablation study presented in Table~\ref{tab:ablation_result} reveals that integrating CAGA with both AGF and CGDM yields superior results compared to any individual module configuration, thereby corroborating the efficacy of our novel attention mechanism in boosting restoration quality.

\section{Conclusion and Future Work}
In this paper, we propose Degradation-Aware Adaptive Context Gating (DACG-IR) for unified image restoration. The proposed framework exploits degradation-aware information from degraded inputs to generate layer-wise and global prompts. The layer prompts guide attention temperature modulation and gated attention responses, while the global prompts enable frequency-domain modulation in deep layers to suppress redundant degradation information. In addition, an adaptive gated fusion (AGF) mechanism is introduced to filter degraded noise in encoder features through spatial and channel gated, guiding the decoder toward high-quality reconstruction. Extensive experiments across diverse restoration scenarios—including all-in-one, single-task, composite, and multi-weather degradations—demonstrate that DACG-IR consistently outperforms existing state-of-the-art methods. While the proposed framework exhibits robust generalization capabilities across a wide spectrum of degradation types, it still lags marginally behind task-specific models in certain domains, notably low-light enhancement and deblurring. Moreover, the Transformer-based design incurs comparatively high computational cost. Our future work will aim at developing more efficient architectures to enhance performance and enable real-world applications.
\section{ACKNOWLEDGEMENT}
This work is supported by the National Natural Science Foundation of China (Nos. 62276218, 12192214 and
62373096), the Fundamental Research Funds for the Central Universities, China (No. 2682024ZTPY055).
\bibliography{DACG-IR-Ref}
\bibliographystyle{IEEEtran}

\newpage
\vfill
\end{document}